%% file: main.tex
\documentclass[letterpaper, 10pt, conference]{ieeeconf}
\IEEEoverridecommandlockouts
\usepackage{amsmath, amssymb, mathtools, nccmath}
\usepackage{array}
\usepackage{environ}
\usepackage{graphicx}
\usepackage{cite}
\usepackage{multirow}
\usepackage{accents}
\usepackage[labelfont=bf]{caption}
\usepackage{subcaption}
\usepackage{booktabs}

\input{macros/macros}

\title{\LARGE \bf
Radar Odometry on $SE(3)$ with Constant Acceleration Motion Prior and Polar Measurement Model
}

\author{Kyle Retan$^{1}$, Frasher Loshaj$^{2}$ and Michael Heizmann$^{3}$
\thanks{Kyle Retan and Frasher Loshaj are with the ZF Group: \tt\small{kyle.retan@zf.com}, \tt\small{frasher.loshaj@zf.com}}%
\thanks{Michael Heizmann is with the Institute for Industrial Information Technology, Karlsruhe Institute of Technology: \tt\small{michael.heizmann@kit.edu}}%
}

\begin{document}

\maketitle
\thispagestyle{empty}
\pagestyle{empty}

\input{abstract/abstract}

\input{introduction/introduction}

\input{related_work/related_work}

\input{problem_definition/problem_definition}

\input{motion_prior/motion_prior}

\input{measurement_model/measurement_model}

\input{data_association/data_association}

\input{evaluation/evaluation}

\input{results/results}

\input{conclusion/conclusion}

\bibliography{literature/bibliography} 
\bibliographystyle{ieeetr}

\end{document}

%% file: macros/macros.tex
\newcommand\numberthis{\addtocounter{equation}{1}\tag{\theequation}}

\newcommand\undervec[1]{\protect\underaccent{\vec}{#1}}
\newcolumntype{L}{>{\centering\arraybackslash}m{1.5cm}}
\newcolumntype{G}{>{\centering\arraybackslash}m{1cm}}

\DeclareMathOperator*{\argmin}{arg\,min}
\DeclareMathOperator*{\argmax}{arg\,max}

%% file: abstract/abstract.tex
\begin{abstract}

This paper presents an approach to radar odometry on $SE(3)$ which utilizes a constant acceleration motion prior. The motion prior is integrated into a sliding window optimization scheme. We use the Magnus expansion to accurately integrate the motion prior while maintaining real-time performance. In addition, we adopt a polar measurement model to better represent  radar detection uncertainties. Our estimator is evaluated using a large real-world dataset from a prototype high-resolution radar sensor. The new motion prior and measurement model signifcantly improve odometry performance relative to the constant velocity motion prior and Cartesian measurement model from our previous work, particularly in roll, pitch and height. 

\end{abstract}

%% file: introduction/introduction.tex
\section{Introduction}

Autonomous vehicles require accurate and robust three dimensional trajectory estimates. Accuracy is achieved by using precise sensors, while robustness requires a variety of sensing technologies that remain functional during challenging operating conditions. Radar-based motion estimation has recently come into focus because radar sensors boast a long range, precise velocity measurements, and are relatively immune to adverse weather. 

The new iteration of automotive radar sensors is capable of measuring in three dimensions, which has motivated the first attempts to estimate radar motion on $SE(3)$. Our previous work in \cite{Retan2021} represented the first attempt to estimate motion on $SE(3)$ with only an automotive radar sensor and a constant velocity motion prior. While the performance for planar motion proved excellent relative to state-of-the-art, the three-dimensional estimates exhibited significant drift in out-of-plane dimensions. We suspected that factors such as sensor bias could play a role, but the choice of measurment and motion model could significantly impact performance, primarily through the data association process.

Recently, motion priors which explicity estimate higher order kinematics have been investigated. The inclusion of acceleration in the vehicle state has been shown to improve performance relative to models which only consider velocity and pose. Given the fact that radar sensors explicitly measure velocity with high precision, we expect the inclusion of acceleration to be especially important for precise motion estimation.

In this work we derive a new constant acceleration motion prior in terms of global pose variables. We combine this new motion prior with a polar coordinate measurement model, which better approximates radar detection uncertainties. We then analyze the impact of the new models on odometry estimation performance relative to our previous work. Together, they significantly improve estimation performance by greatly reducing estimation drift in out-of-plane dimensions and by increasing estimator stability in challenging driving scenarios.

%% file: related_work/related_work.tex
\section{Related Work}
\subsection{Radar Odometry}
Traditionally, radar odometry has been performed using automotive radar sensors, which output a point cloud representation of the environment with point-wise radial velocity measurements. Odometry can be estimated using various combinations of spatial and radial velocity measurements together with various assumptions regarding vehicle motion. In \cite{Kellner2013}, non-holonomic motion is assumed in order to estimate planar odometry from angular and radial velocity measurements. A second radar sensor was included in \cite{Kellner2014} to estimate planar motion without motion constraints. The incorporation of spatial measurements in \cite{Barjenbruch2015} and \cite{Rapp2015} increased accuracy and made full planar motion estimates from a single radar sensor possible. A decoupled linear and angular velocity estimate was introduced in \cite{8995552}. All of these approaches were restricted to planar motion. Recently, Doer et al. proposed a fusion of an FMCW radar and IMU in \cite{Doer2020} and \cite{Doer2021} for 3D odometry estimation with online calibration, which demonstrates a growing interest in expanding radar odometry estimates to three dimensions.%
\subsection{Motion Priors}

Estimating the trajectory of a robot $\mathbf{T}(t) \in SE(3)$ is one of the core robotics tasks. In recent years, interest in higher order kinematics such as velocity and acceleration has grown. In \cite{Anderson2015} a white-noise-on-acceleration (WNOA) or constant velocity prior for a relative formulation of bundle adjustment was developed. An absolute formulation of the constant velocity model was then proposed in \cite{Barfoot2017}, which was also used as the motion prior in our previous radar odometry work in \cite{Retan2021}. Although planar motion estimates were very good, the estimates of altitude, pitch and roll exhibited a large drift. Tang et al. demonstrated in \cite{Tang2018} that the inclusion of acceleration in the robot state could significantly enhance LIDAR odometry esimates relative to constant velocity motion priors. More specifically, it was found to improve drift in out-of-plane dimensions -- precisely the weak point of our previous work. In \cite{8968328} Wong et al. present a data-driven approach to learning a parameterizable family of motion priors from ground truth data. 

We were inspired by the improvements observed from the inclusion of higher order kinematics in the robot state. However, we find the use of absolute pose and and body frame velocity in \cite{Barfoot2017} to be more intuitive. We therefore develop a constant acceleration model using \textit{absolute} pose variables and body-frame kinematics.

%% file: problem_definition/problem_definition.tex
\section{Problem Definition}
Here we present our approach to estimating motion on $SE(3)$ using radar scans. We are given a prior state $\check{\mathbf{x}}_k$, $K$ consecutive radar scans $\boldsymbol{\mathcal{Z}}_{1:K}$ and a map of our environment $\boldsymbol{\mathcal{M}}$ consisting of homogeneous points. We wish to estimate a trajectory of states $\hat{\mathbf{x}}_{0:k}$, which we can formulate as the maximum a posteriori (MAP) estimation problem
\begin{align}
\hat{\mathbf{x}}_{0:K} &= \argmax_{\mathbf{x}_{0:K}} p\left( \mathbf{x}_0 \vert \check{\mathbf{x}}_0 \right) \prod_{k=1}^K p\left( \mathbf{x}_k \vert \mathbf{x}_{k-1} \right) \prod p\left(\boldsymbol{\mathcal{Z}}_k \vert \mathbf{x}_k \right)
\end{align}
by using the Markovian property of states and the assumption of conditional measurement independence. The negative log-likelihood cost function 
\begin{equation}
J = J_{v,0} + \sum_{k=1}^K J_{v,k} + \sum_{k=1}^K J_{z,k} \label{eq:cost_function}
\end{equation}
is then minimized to obtain a solution. Here $J_{v,0}$, $J_{v,k}$ and $J_{z,k}$ denote the prior, motion prior and measurement cost terms respectively. In this work, we first derive a constant acceleration motion prior for $SE(3)$ using abolute poses and body-frame kinematics. We then expand and modify our previous measurement model for polar radar measurements. Next, we analzye the effect of the improved motion and measurement models on the data association task. Finally, the proposed estimator is implemented and tested on a large real-world dataset from a high resolution radar sensor.
\begin{figure*}
\vspace*{0.15cm}
     \centering
     \begin{subfigure}[t]{0.4\textwidth}
         \centering
         \includegraphics[width=\textwidth]{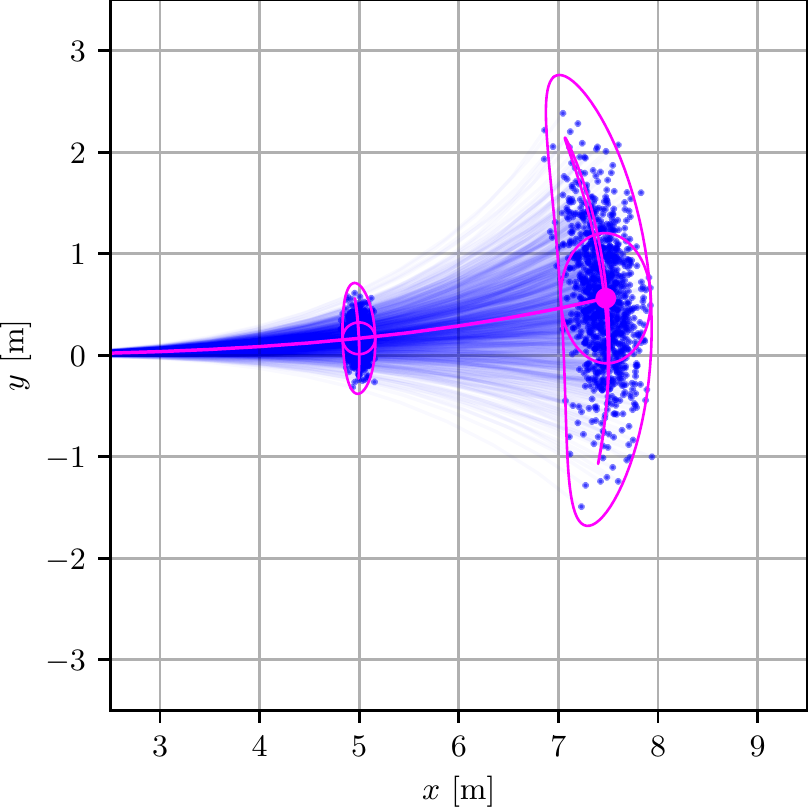}
     \end{subfigure}
     \hspace{7em}%
     \begin{subfigure}[t]{0.4\textwidth}
         \centering
         \includegraphics[width=\textwidth]{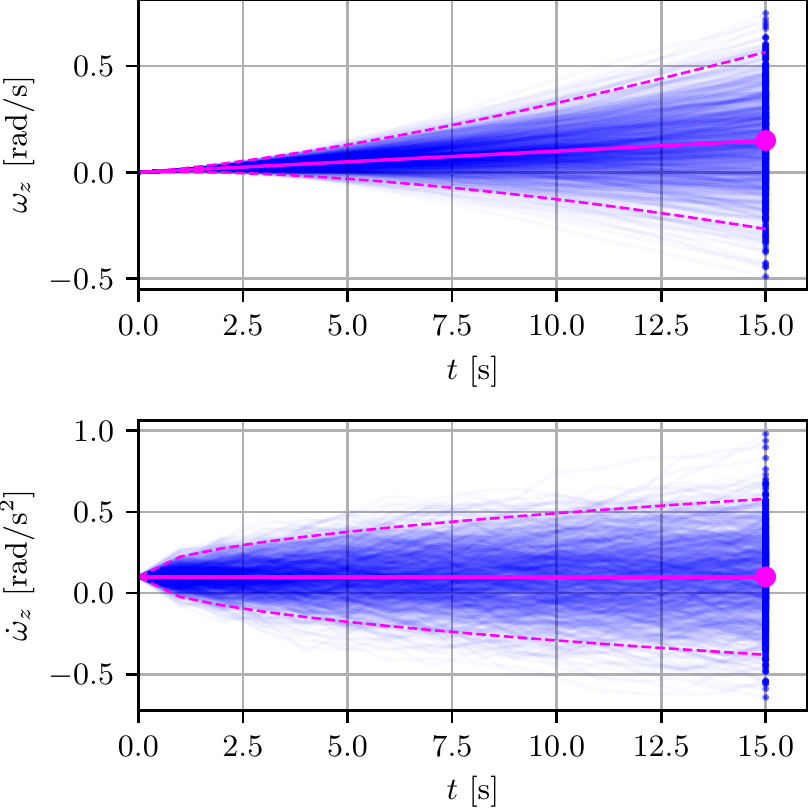}
     \end{subfigure}
     \caption{Simulation showing position (left), velocity and acceleration (right) of constant acceleration motion prior for pure yaw acceleration, meaning $		     \dot{\boldsymbol{\omega}} = [0 \; 0 \; \dot{\omega_z}]^T$. Mean and 3-$\sigma$ covariance ellipses are depiced in pink. Monte Carlo samples from the prior are shown in blue.}
     \label{fig:motion_prior}
\end{figure*}

%% file: motion_prior/motion_prior.tex
\section{Motion Prior}
In this section we first introduce the random variable utilized in our motion prior. We then present the stochastic differential equation (SDE) used to describe $SE(3)$ kinematics with acceleration. Finally, we discretize these equations to obtain the state and covariance transitions necessary for our optimization problem. 

\subsection{State Variable Definition}
We begin by defining our state variable $\mathbf{x}$ as
\begin{equation}
\mathbf{x} = \left\{ \mathbf{T}, \boldsymbol{\varpi}, \dot{\boldsymbol{\varpi}} \right\},
\end{equation}
where $\mathbf{T} \in SE(3)$ is the vehicle pose with vector space representation $\boldsymbol{\xi} = [ \boldsymbol{\rho}^T \; \boldsymbol{\phi}^T ]^T \in \mathbb{R}^{6}$. The rotational and translational components are given by $\boldsymbol{\rho}, \boldsymbol{\phi} \in \mathbb{R}^{3}$ respectively. The vector $\boldsymbol{\varpi} = [ \boldsymbol{\nu}^T \; \boldsymbol{\omega}^T ]^T \in \mathbb{R}^6$ represents the body frame velocity with linear and angular components $\boldsymbol{\nu}, \boldsymbol{\omega} \in \mathbb{R}^{3}$. Similarly, the vector $\dot{\boldsymbol{\varpi}} = [ \dot{\boldsymbol{\nu}}^T, \dot{\boldsymbol{\omega}}^T ]^T$ denotes the body frame acceleration with linear and angular components $\dot{\boldsymbol{\nu}}, \dot{\boldsymbol{\omega}} \in \mathbb{R}^3$. We use the overloaded operator $(\cdot)^\wedge$ to transform vectors $\boldsymbol{\phi} \in \mathbb{R}^{3}$ and $\boldsymbol{\xi} \in \mathbb{R}^{6}$ into members of the Lie algebras $\mathfrak{so}(3)$ and $\mathfrak{se}(3)$ respectively.

The random variable $\mathbf{x}$ has the form
\begin{align}
	\mathbf{x} &= \boldsymbol{\gamma} \oplus \check{\mathbf{x}} = \left\{\exp \left( \boldsymbol{\xi}^\wedge \right) \check{\mathbf{T}}, \check{\boldsymbol{\varpi}} + \boldsymbol{\eta}, \check{\dot{\boldsymbol{\varpi}}} + \boldsymbol{\zeta} \right\} \\
 	\boldsymbol{\gamma} &= 
	\begin{bmatrix}
		\boldsymbol{\xi} \\
		\boldsymbol{\eta} \\
		\boldsymbol{\zeta} \\
	\end{bmatrix} \sim \mathcal{N}\left( \mathbf{0}, \mathbf{P} \right), \;\; \mathbf{P} \in \mathbb{R}^{18 \times 18}
\end{align}
where $\check{\mathbf{T}}$, $\check{\boldsymbol{\varpi}}$ and $\check{\dot{\boldsymbol{\varpi}}}$ refer to the large,  deterministic components of pose, velocity and acceleration, while $\boldsymbol{\xi},\boldsymbol{\eta}, \boldsymbol{\zeta}\in \mathbb{R}^6$ denote small perturbations around these values.

The resulting distribution can be understood as a classical Gaussian random variable $\boldsymbol{\gamma}$ which induces a distribution on the composite manifold $SE(3) \times \mathbb{R}^{12}$
\begin{align}
1 &= \int_{\mathbb{R}^{12}} \frac{1}{\sqrt{2 \pi \det(\mathbf{P})}} \exp \left( -\frac{1}{2} \gamma ^T \mathbf{P}^{-1} \gamma \right) \mathrm{d}\boldsymbol{\gamma} \\
&= \int_{SE(3) \times \mathbb{R}^{12}} \frac{1}{\sqrt{ (2\pi)^{12} \det(\mathbf{P})}}\\
& \;\;\;\; \times \exp \left(-\frac{1}{2} (\mathbf{x} \ominus \check{\mathbf{x}})^T \mathbf{P}^{-1} (\mathbf{x} \ominus \check{\mathbf{x}}) \right) \frac{1}{\det{\boldsymbol({\Psi})}} \mathrm{d}\mathbf{x},
\end{align}
where we use the change of coordinates and corresponding differentials
\begin{align}
\boldsymbol{\gamma} &= \mathbf{x} \ominus \check{\mathbf{x}} \\
\mathrm{d}\mathbf{x} &= \left| \det(\boldsymbol{\Psi}) \right| \mathrm{d}\boldsymbol{\gamma} \\
\boldsymbol{\Psi} &= \frac{\partial \left( \mathbf{x} \ominus \check{\mathbf{x}} \right)}{\partial \boldsymbol{\gamma}} =
\begin{bmatrix}
\boldsymbol{\mathcal{J}}(\boldsymbol{\xi}) & \mathbf{0} & \mathbf{0} \\
\mathbf{0} & \mathbf{1} & \mathbf{0} \\
\mathbf{0} & \mathbf{0} & \mathbf{1}
\end{bmatrix}.
\end{align}
Here $\boldsymbol{\mathcal{J}} \in \mathbb{R}^{6 \times 6}$ denotes the left Jacobian of $SE(3)$. We introduce the operatorand $\ominus$ for convenience, which we define as 
\begin{align}
\mathbf{x}_1 \ominus \mathbf{x}_2 &=
\begin{bmatrix}
\ln\left(\mathbf{T}_1 \mathbf{T}_2 ^{-1} \right) \\
\boldsymbol{\varpi}_1 - \boldsymbol{\varpi}_2 \\
\dot{\boldsymbol{\varpi}}_1 - \dot{\boldsymbol{\varpi}}_2
\end{bmatrix}.
\end{align}
This distribution forms the basis of our motion model, which is also an integral component of our optimization problem as well as our data association strategy.
\subsection{Kinematics}
Our constant acceleration motion prior is a nonlinear SDE of the form
\begin{align}
\dot{\mathbf{T}}(t) &= \boldsymbol{\varpi}(t)^\wedge \mathbf{T}(t) \label{eq:pose_derivative} \\
\ddot{\boldsymbol{\varpi}}(t) &= \mathbf{w}(t), \; \mathbf{w}(t) \sim \mathcal{N}\left( \mathbf{0}, \mathbf{Q} \right), \label{eq:acceleration_derivative}
\end{align}
which can be separated into nominal and perturbation components using the $SE(3)$ constraint-sensitive linearization techniques outlined in \cite{Barfoot2017}. Fig. \ref{fig:motion_prior} shows simulation results of the motion prior for pure yaw acceleration. Our kinematics model then consists of a nonlinear nominal component which captures the evolution of the mean of our SDE
\begin{align}
\check{\dot{\mathbf{T}}}(t) &= \check{\boldsymbol{\varpi}}(t)^\wedge \check{\mathbf{T}}(t) \label{eq:pose_derivative} \\
\check{\ddot{\boldsymbol{\varpi}}}(t) &= \mathbf{0},
\end{align}
and a linear time-variant (LTV) component which captures the evolution of its covariance:
\begin{align}
	\underbrace{
		\begin{bmatrix}
			\dot{\boldsymbol{\xi}}(t) \\ 
			\dot{\boldsymbol{\eta}}(t) \\
			\dot{\boldsymbol{\zeta}}(t)
		\end{bmatrix}
	}_{\dot{\boldsymbol{\gamma}}(t)} 
	&=
	\underbrace{
		\begin{bmatrix}
			\boldsymbol{\varpi}^{\curlywedge}(t) & \mathbf{1} & \mathbf{0} \\
			\mathbf{0} & \mathbf{0} & \mathbf{1} \\
			\mathbf{0} & \mathbf{0} & \mathbf{0}
		\end{bmatrix}
	}_{\mathbf{A}}
	\underbrace{
		\begin{bmatrix}
			\boldsymbol{\xi}(t) \\
			\boldsymbol{\eta}(t) \\
			\boldsymbol{\zeta}(t)	
		\end{bmatrix}
	}_{\boldsymbol{\gamma}(t)} +
	\underbrace{
		\begin{bmatrix}
			\mathbf{0} \\
			\mathbf{0} \\
			\mathbf{1}
		\end{bmatrix}
	}_{\mathbf{L}}
	\mathbf{w}(t) \label{eq:pertrubation_diff_eq}\\
	\mathbf{w}(t) &\sim \mathcal{N}(\mathbf{0}, \mathbf{Q}).
\end{align}
The state perturbation $\boldsymbol{\gamma}(t)$ consists of the pose, velocity and acceleration perturbations $\boldsymbol{\xi}(t), \boldsymbol{\eta}(t), \boldsymbol{\zeta}(t) \in \mathbb{R}^{6}$. The operator $(\cdot)^{\curlywedge}$ transforms the vector $\boldsymbol{\varpi}$ to its adjoint representation, given by
\begin{equation}
\boldsymbol{\varpi}^{\curlywedge} =
\begin{bmatrix}
\boldsymbol{\nu} \\
\boldsymbol{\omega}
\end{bmatrix} =
\begin{bmatrix}
\boldsymbol{\omega}^\wedge & \boldsymbol{\nu}^\wedge \\
\mathbf{0} & \boldsymbol{\omega}^\wedge
\end{bmatrix} \in \mathbb{R}^{6 \times 6}.
\end{equation}
We must discretize the motion prior via integration over the interval $t \in \left[ t_{k-1}, t_k \right]$, which represents the time between two consecutive radar measurements. The discretized quantities can then be used for optimization to solve (\ref{eq:cost_function}). In the following, we present our discretization approach.
\subsubsection{Nominal Solution}
We can obtain a solution for $\boldsymbol{\varpi}(t)$ and $\dot{\boldsymbol{\varpi}}(t)$ via straigtforward integration, which yields
\begin{align}
\dot{\boldsymbol{\varpi}}(t) &= \dot{\boldsymbol{\varpi}}(t_0)\\
\boldsymbol{\varpi}(t) &= \boldsymbol{\varpi}(t_0) + (t - t_0)\dot{\boldsymbol{\varpi}}(t_0).
\end{align}
Unfortunately, as the velocity $\boldsymbol{\varpi}(t)$ is now time-dependent, the solution to (\ref{eq:pose_derivative}) is no longer simply the integral
\begin{equation}
\mathbf{T}(t) = \exp \left( \int_{t_0}^{t}  \boldsymbol{\varpi}^\wedge (\tau)  \mathrm{d}\tau\right) \cdot \mathbf{T}(t_0)
\end{equation}
because the velocity matrix $\boldsymbol{\varpi}^\wedge(t)$, does not commute with itself at different times, meaning $\left[\boldsymbol{\varpi}^\wedge(t_1), \boldsymbol{\varpi}^\wedge(t_2) \right] \neq \mathbf{0}$ for $t_1, t_2 \in \left[t_0, t \right]$.
Various techniques exist for solving (\ref{eq:pose_derivative}) numerically, such as Runge-Kutta methods adapted for $SE(3)$. However, these methods can be cumbersome and potentially too slow for real-time applications. The selected method must be both computationally inexpensive and accurate over time intervals between radar scans. We choose an approach similar to \cite{Huber2019}, which uses the Magnus expansion method outline in \cite{Blanes2009} to approximate the integral's solution as the matrix exponential
\begin{equation}
\mathbf{T}(t) = \exp(\boldsymbol{\mathcal{S}_{\boldsymbol{\varpi}}}(t,t_0)) \mathbf{T}(t_0),
\end{equation}
where $\boldsymbol{\mathcal{S}_{\boldsymbol{\varpi}}}(t,t_0) \in \mathfrak{se}(3)$ is constructed via the series expansion
\begin{equation}
\boldsymbol{\mathcal{S}}_{\boldsymbol{\varpi}}(t,t_0) = \sum_{n=1}^\infty \boldsymbol{\mathcal{S}}_{\boldsymbol{\varpi}}^{(n)}(t,t_0),
\end{equation}
in which the first three terms are given by
\begin{flalign*}
& \boldsymbol{\mathcal{S}}_{\boldsymbol{\varpi}}^{(1)}(t,t_0) = \int_{t_0}^t \boldsymbol{\varpi}^\wedge (t_1) \mathrm{d}t_1 \numberthis & \\
& \boldsymbol{\mathcal{S}}_{\boldsymbol{\varpi}}^{(2)}(t,t_0) = \frac{1}{2} \int_{t_0}^t \mathrm{d}t_1 \int_{t_0}^{t_1} \mathrm{d}t_2 \left\{ \left[ \boldsymbol{\varpi}^\wedge (t_1), \boldsymbol{\varpi}^\wedge (t_2)
 \right] \right\} \numberthis & \\
& \boldsymbol{\mathcal{S}}_{\boldsymbol{\varpi}}^{(3)}(t,t_0) = \frac{1}{6} \int_{t_0}^t \mathrm{d}t_1 \int_{t_0}^{t_1} \mathrm{d}t_2 \int_{t_0}^{t_2} \mathrm{d}t_3 \numberthis &\\ 
&\;\;\; \qquad \qquad \;\;\, \big\{ \big[ \left[ \boldsymbol{\varpi}^\wedge (t_1), \boldsymbol{\varpi}^\wedge (t_2)
 \right], \boldsymbol{\varpi}^\wedge(t_3) \big] & \\ 
&\;\;\;\;\;\; \qquad \qquad + \big[ \boldsymbol{\varpi}^\wedge (t_1), \left[ \boldsymbol{\varpi}^\wedge (t_2), \boldsymbol{\varpi}^\wedge(t_3) \right] \big] \big\}.&
\end{flalign*}
We then evaluate the expansion over the interval ${t \in \left[t_{k-1}, t_k \right]}$ to obtain a relatively compact expression for each term:
\begin{align}
\boldsymbol{\mathcal{S}}_{\boldsymbol{\varpi}}^{(1)}(\Delta t) &= \Delta t \boldsymbol{\varpi}_{k-1}^\wedge + \frac{\Delta t^2}{2} \dot{\boldsymbol{\varpi}}_{k-1}^\wedge \\
\boldsymbol{\mathcal{S}}_{\boldsymbol{\varpi}}^{(2)}(\Delta t) &= \frac{\Delta t ^3}{12} [ \dot{\boldsymbol{\varpi}}_{k-1}^\wedge, \boldsymbol{\varpi}_{k-1}^\wedge] \\
\boldsymbol{\mathcal{S}}_{\boldsymbol{\varpi}}^{(3)}(\Delta t) &= \frac{\Delta t^5}{240} \big[\dot{\boldsymbol{\varpi}}_{k-1}^\wedge [ \dot{\boldsymbol{\varpi}}_{k-1}^\wedge, \boldsymbol{\varpi}_{k-1}^\wedge] \big], 
\end{align}
where we use the notation $\boldsymbol{\mathcal{S}}_{\boldsymbol{\varpi}}(\Delta t) = \boldsymbol{\mathcal{S}}_{\boldsymbol{\varpi}}(t_{k}, t_{k-1})$ and $\Delta t = t_k - t_{k-1}$. We also use the index $k$ to refer to a variable evaluated at time $t_k$. This series expansion is inexpensive to evaluate and very accurate over small time intervals. The integral can be approximated using the standard exponential map, which ensures that ${\mathbf{T}_k \in SE(3)}$. It is also clear that the approximation simplifies to the constant velocity case $\boldsymbol{\mathcal{S}}_{\boldsymbol{\varpi}}(\Delta t) = \Delta t\boldsymbol{\varpi}_{k-1}^\wedge$ for $\dot{\boldsymbol{\varpi}}_{k-1}^\wedge = \mathbf{0}$. We can now write the discrete-time transition of the mean over the interval ${t \in \left[t_{k-1}, t_k \right]}$ as 
\begin{align}
\mathbf{x}_k &= \mathbf{f}\left( \mathbf{x}_{k-1} \right) = \{\mathbf{T}_k, \boldsymbol{\mathcal{\varpi}}_k, \dot{\boldsymbol{\mathcal{\varpi}}}_k \} \\
\mathbf{T}_k &= \exp\left( \boldsymbol{\mathcal{S}}_{\boldsymbol{\varpi}}(\Delta t) \right) \mathbf{T}_{k-1}\\
\boldsymbol{\varpi}_k &= \boldsymbol{\varpi}_{k-1} + \Delta t \dot{\boldsymbol{\varpi}}_{k-1} \\ 
\dot{\boldsymbol{\varpi}}_k &=  \dot{\boldsymbol{\varpi}}_{k-1}.
\end{align}
\subsection{Perturbation Solution}
To discretize our perturbation equations, which capture the evolution of the covariance of our state variable $\mathbf{P}(t)$, we must integrate the linear time-varying SDE \cite{Srkk2019}
\begin{align}
\dot{\boldsymbol{\gamma}}(t) &= \mathbf{A}(t) \boldsymbol{\gamma}(t) + \mathbf{L} \mathbf{w}(t) \\
\dot{\mathbf{P}}(t) &= \mathbf{A}(t)\mathbf{P}(t)+\mathbf{P}(t)\mathbf{A}(t)^{T}+\mathbf{L}\mathbf{Q}\mathbf{L}^{T}, \label{eq:covariance_diff_eq}
\end{align}
whose general solution is given by
\begin{align*}
\boldsymbol{\gamma}(t) &= \boldsymbol{\Phi}(t,t_0)\boldsymbol{\gamma}(t_0) \numberthis \\
\mathbf{P}(t) &= \boldsymbol{\Phi}(t,t_0)\mathbf{P}(t_0)\boldsymbol{\Phi}(t,t_0)^T \\
&+  \int_{t_0}^{t} \boldsymbol{\Phi}(t, \tau) \mathbf{L} \mathbf{Q} \mathbf{L}^T \boldsymbol{\Phi}(t, \tau)^T \mathrm{d}\tau, \numberthis
\end{align*}
where $\boldsymbol{\Phi}\left(t, s \right)$ denotes the state transition matrix. The state transition matrix itself is
\begin{equation}
\boldsymbol{\Phi}(t,t_0) = \boldsymbol{\Upsilon}(t)\boldsymbol{\Upsilon}(t_0)^{-1},
\end{equation}
where $\boldsymbol{\Upsilon}(t)$ denotes the fundamental solution to the linear differential equation
\begin{equation}
\dot{\boldsymbol{\Upsilon}}(t) = \mathbf{A}(t)\boldsymbol{\Upsilon}(t), \;\;\; \boldsymbol{\Upsilon}(0) = \mathbf{1}.
\end{equation}
Discretization involves the computation of the discrete state transition $\mathbf{F}_k$ and the process noise covariance $\mathbf{Q}_k$, given by
\begin{align}
\mathbf{F}_k &= \boldsymbol{\Phi}\left( t_k, t_{k-1} \right) \\
\mathbf{Q}_k &= \int_{t_{k-1}}^{t_k} \boldsymbol{\Phi}\left(t_k, s \right) \mathbf{L} \mathbf{Q} \mathbf{L}^T \boldsymbol{\Phi}\left(t_k, s \right)^T, \label{eq:process_noise_covariance_integral}
\end{align}
Alternatively, we can solve for $\mathbf{F}_k$ and $\mathbf{Q}_k$ using a matrix fraction decomposition. For example, in \cite{Srkk2019} $\mathbf{P}(t)$ solves (\ref{eq:covariance_diff_eq}) if the matrices $\mathbf{C}(t)$ and $\mathbf{D}(t)$ solve the homogeneous differential equation
\begin{equation}
\underbrace{
\begin{bmatrix}
\dot{\mathbf{C}}(t) \\
\dot{\mathbf{D}}(t)
\end{bmatrix}}_{\dot{\mathbf{Y}}(t)}
=
\underbrace{
\begin{bmatrix}
\mathbf{A}(t) & \mathbf{L}\mathbf{Q}\mathbf{L}^T \\
\mathbf{0} & -\mathbf{A}^T(t)
\end{bmatrix}
}_{\mathbf{M}(t)}
\underbrace{
\begin{bmatrix}
\mathbf{C}(t) \\
\mathbf{D}(t)
\end{bmatrix}}_{\mathbf{Y}(t)} \label{eq:matrix_fraction}
\end{equation}
and $\mathbf{P}\left(t \right) = \mathbf{C}\left(t \right) \mathbf{D}^{-1}\left(t \right)$. We note that (\ref{eq:matrix_fraction}) is also a linear time-varying differential equation, which allows us to again use the Magnus expansion to obtain a solution. The Magnus expansion of our system $\mathbf{M}(t)$ can be simplified to
\begin{align}
\boldsymbol{\mathcal{S}}_{\mathbf{M}}(t, t_0) &= 
\begin{bmatrix}
\boldsymbol{\mathcal{S}}_\mathbf{A}(t, t_0) & \boldsymbol{\mathcal{G}}(t,t_0)  \\
\mathbf{0} & -\boldsymbol{\mathcal{S}}_\mathbf{A}(t, t_0)^{T}
\end{bmatrix},
\end{align}
where $\boldsymbol{\mathcal{S}}_\mathbf{A}(t, t_0)$ denotes the Magnus expansion of $\mathbf{A}(t)$ and $\boldsymbol{\mathcal{G}}(t, t_0)$ consist of terms coupling $\mathbf{A}(t)$ and $\mathbf{L}\mathbf{Q}\mathbf{L}^T$, which are zero up to expansion order three. The solution to (\ref{eq:matrix_fraction}) is found via exponentiation of the resulting expansion, yielding
\begin{align}
\mathbf{Y}(t)
&=
\exp \left(
\boldsymbol{\mathcal{S}}_{\mathbf{M}}(t, t_0)
\right)
\mathbf{Y}(t_0) \label{ex:matrix_fraction_exponential} \\
&=
\begin{bmatrix}
\exp(\boldsymbol{\mathcal{S}}_{\mathbf{A}}(t, t_0)) & \mathbf{G}(t, t_0) \\
\mathbf{0} & \exp(\boldsymbol{\mathcal{S}}_{\mathbf{A}}(t, t_0))^{-T}
\end{bmatrix}
\mathbf{Y}(t_0). 
\end{align}
The state transition matrix ${\boldsymbol{\Phi}(t, t_0) = \exp\left(\boldsymbol{\mathcal{S}}_{\mathbf{A}}(t, t_0)\right)}$ occupies the upper left position. If we evaluate this expression over the interval ${t \in \left[t_{k-1}, t_k \right]}$ and choose $\mathbf{C}(t_{k - 1}) = \mathbf{0}$, ${\mathbf{D}(t_{k-1}) = \mathbf{1}}$, we obtain the expressions for the discrete time state transition $\mathbf{F}_k$ and process noise covariance $\mathbf{Q}_k$
\begin{align}
\mathbf{F}_k &= \exp(\boldsymbol{\mathcal{S}}_{\mathbf{A}}(\Delta t)) \\
\mathbf{Q}_k &= \mathbf{G}(\Delta t) \cdot \exp(\boldsymbol{\mathcal{S}}_{\mathbf{A}}(\Delta t))^T.
\end{align}
which allows us to construct cost terms of the form
\begin{align}
J_{v,k} &= \mathbf{e}_{v,k}^T \mathbf{Q}_k^{-1} \mathbf{e}_{v,k} \\
\mathbf{e}_{v,k} &= \boldsymbol{\mathbf{F}}_k \gamma_{k-1} - \boldsymbol{\gamma}_k.
\end{align}
To summarize, we discretize the SDE describing the evolution of the perturbation $\boldsymbol{\gamma}(t)$ and its covariance $\mathbf{P}(t)$ by computing the matrix exponential of the magnus expansion $\boldsymbol{\mathcal{S}}_{\mathbf{M}}(\Delta t)$. Numerical methods such as the Pad\'e approximant can be used to compute the exponential efficiently. The resulting state transition and process noise covariance matrices can then be used to solve (\ref{eq:cost_function}) numerically.
\subsection{Magnus Expansion of $\boldsymbol{\mathcal{S}}_{\mathbf{A}}$}
We now turn to the expansion $\boldsymbol{\mathcal{S}}_{\mathbf{A}}(\Delta t)$, which is required to compute the solution to (\ref{eq:matrix_fraction}). Due to the structure of our problem, the expansion will have the general form
\begin{equation}
\boldsymbol{\mathcal{S}}_{\mathbf{A}}(\Delta t) = 
\begin{bmatrix}
\boldsymbol{\mathcal{S}}_{\mathbf{A}, 11}(\Delta t) & \boldsymbol{\mathcal{S}}_{\mathbf{A}, 12}(\Delta t) & \boldsymbol{\mathcal{S}}_{\mathbf{A}, 13}(\Delta t) \\
\mathbf{0} & \mathbf{0} & \Delta t\mathbf{1} \\
\mathbf{0} & \mathbf{0} & \mathbf{0}
\end{bmatrix}
\end{equation}
for an arbitrary expansion order. If we again use a third order expansion, we obtain the terms
\begin{align*}
\boldsymbol{\mathcal{S}}_{\mathbf{A},11}^{(1)}(\Delta t) &= \Delta t\boldsymbol{\varpi}_{k-1}^{\curlywedge}+\frac{\Delta t^{2}}{2}\dot{\boldsymbol{\varpi}}_{k-1}^\curlywedge \numberthis \\
\boldsymbol{\mathcal{S}}_{\mathbf{A},11}^{(2)}(\Delta t) &= \frac{\Delta t^{3}}{12} \left[ \dot{\boldsymbol{\varpi}}_{k-1}^\curlywedge, \boldsymbol{\varpi}_{k-1}^\curlywedge \right] \numberthis \\
\boldsymbol{\mathcal{S}}_{\mathbf{A},11}^{(3)}(\Delta t) &=\frac{\Delta t^5}{240}\left[ \dot{\boldsymbol{\varpi}}_{k-1}^\curlywedge, \left[ \dot{\boldsymbol{\varpi}}_{k-1}^\curlywedge, \boldsymbol{\varpi}_{k-1}^\curlywedge \right] \right] \numberthis \\ \\
\boldsymbol{\mathcal{S}}_{\mathbf{A},12}^{(1)}(\Delta t) &= \Delta t\mathbf{1} \numberthis \\
\boldsymbol{\mathcal{S}}_{\mathbf{A},12}^{(2)}(\Delta t) &= \frac{\Delta t^3}{12}\dot{\boldsymbol{\varpi}}^\curlywedge_{k-1} \numberthis \\
\boldsymbol{\mathcal{S}}_{\mathbf{A},12}^{(3)}(\Delta t) &= \frac{\Delta t^5}{240}\left( \dot{\boldsymbol{\varpi}}^\curlywedge_{k-1} \right)^2. \numberthis
\end{align*}
The element $\boldsymbol{\mathcal{S}}_{\mathbf{A},13}(\Delta t)$ is zero up to third order.
With the expansion terms in hand, we can now numerically evaluate the matrix exponential in (\ref{ex:matrix_fraction_exponential}) to obtain approximations for $\mathbf{F}_k$ and $\mathbf{Q}_k$.
\subsection{Closed-Form State Transition Exponential}
It is also possible to represent the matrix exponential ${\mathbf{F}_k = \exp(\boldsymbol{\mathcal{S}}_{\mathbf{A}}(\Delta t))}$ in closed form. We can write $\mathbf{F}_k$ as the series
\begin{align}
\mathbf{F}_k &= \sum_{n=0}^{\infty} \frac{\boldsymbol{\mathcal{S}}_{\mathbf{A}}(\Delta t)^n}{n!} \\
&= 
\begin{bmatrix}
\mathbf{F}_{k,11}(\Delta t) & \mathbf{F}_{k,12}(\Delta t) & \mathbf{F}_{k,13}(\Delta t) \\
\mathbf{0} & \mathbf{1} & \Delta t\mathbf{1} \\
\mathbf{0} & \mathbf{0} & \mathbf{1}
\end{bmatrix} \\
\end{align}
with elements
\begin{align}
\mathbf{F}_{k,11}(\Delta t) &= \sum\limits^\infty_{n=0} \dfrac{\boldsymbol{\mathcal{S}}_{\mathbf{A},11}(\Delta t)^n}{n!} \\
&= \exp \left(\boldsymbol{\mathcal{S}}_{\mathbf{A},11}(\Delta t)\right) \\
\mathbf{F}_{k,12}(\Delta t) &= \sum\limits^\infty_{n=0} \dfrac{\boldsymbol{\mathcal{S}}_{\mathbf{A},11}(\Delta t)^n}{(n+1)!} \cdot \boldsymbol{\mathcal{S}}_{\mathbf{A},12}(\Delta t) \\
&= \boldsymbol{\mathcal{J}}(\boldsymbol{\mathcal{S}}_{\mathbf{A},11}(\Delta t))\cdot \boldsymbol{\mathcal{S}}_{\mathbf{A},12}(\Delta t) \\
\mathbf{F}_{k,13}(\Delta t) &= \sum\limits^\infty_{n=0} \Bigg\{ \dfrac{\boldsymbol{\mathcal{S}}_{\mathbf{A},11}(\Delta t)}{(n+1)!} \cdot \boldsymbol{\mathcal{S}}_{\mathbf{A},13}(\Delta t) \\
&\;\;\;\;\;\;\;\;\;+ \dfrac{\boldsymbol{\mathcal{S}}_{\mathbf{A},11}(\Delta t)}{(n+2)!} \cdot \boldsymbol{\mathcal{•}{S}}_{\mathbf{A},12}(\Delta t)\Bigg\} \\
&= \boldsymbol{\mathcal{J}}(\boldsymbol{\mathcal{S}}_{\mathbf{A},11}(\Delta t)) \cdot \boldsymbol{\mathcal{S}}_{\mathbf{A},13}(\Delta t)  \\ 
&\;\;\;\;\;\;\;\;\; + \Delta t \boldsymbol{\mathcal{H}}(\boldsymbol{\mathcal{S}}_{\mathbf{A},11}(\Delta t)) \cdot \boldsymbol{\mathcal{S}}_{\mathbf{A},12}(\Delta t).
\end{align}
This matrix is valid for an arbitrary Magnus expansion order. We use a third order Magnus expansion, for which  $\boldsymbol{\mathcal{S}}_{\mathbf{A},13}(\Delta t) = \mathbf{0}$. In this case $\mathbf{F}_{k,13}(\Delta t)$ simplifies to
\begin{equation}
\mathbf{F}_{k,13}(\Delta t) = \Delta t \boldsymbol{\mathcal{H}}(\boldsymbol{\mathcal{S}}_{\mathbf{A},11}(\Delta t)) \cdot \boldsymbol{\mathcal{S}}_{\mathbf{A},12}(\Delta t).
\end{equation}
We can find a closed form expression for $\boldsymbol{\mathcal{H}}\left(\cdot\right)$ by using the identity
\begin{equation}
\left(\boldsymbol{\xi}^{\curlywedge}\right)^{5}+2\phi^{2}\left(\boldsymbol{\xi}^{\curlywedge}\right)^{3}+\phi^{4}\boldsymbol{\xi}^{\curlywedge} \equiv \mathbf 0,
\end{equation}
with $\phi = \left\Vert \boldsymbol{\phi} \right\Vert$, which allows us to express quintic and higher order terms in terms of lower order terms, giving us
\begin{equation}
\boldsymbol{\xi}^\curlywedge = \frac{1}{2!}+\alpha_{1}\left(\boldsymbol{\xi}^{\curlywedge}\right)+\alpha_{2}\left(\boldsymbol{\xi}^{\curlywedge}\right)^{2}+\alpha_{3}\left(\boldsymbol{\xi}^{\curlywedge}\right)^{3}+\alpha_{2}\left(\boldsymbol{\xi}^{\curlywedge}\right)^{4},
\end{equation}
with
\begin{align}
\alpha_{1} &= \frac{4\phi+\phi\cos\phi-5\sin\phi}{2\phi^{3}} \\
\alpha_{2} &= \frac{2\phi^{2}+\phi\sin\phi+6\cos\phi-6}{2\phi^{4}} \\
\alpha_{3} &= \frac{2\phi+\phi\cos\phi-3\sin\phi}{2\phi^{5}} \\
\alpha_{4} &= \frac{\phi^{2}+\phi\sin\phi+4\cos\phi-4}{2\phi^{6}}.
\end{align}

The closed form solution may be preferred for applications in which an alternative method is used to numerically integrate or approximate the process noise covariance $\mathbf{Q}_k$.
\begin{figure}[t]
\vspace*{0.15cm}
\centering
\includegraphics[width=0.4\textwidth]{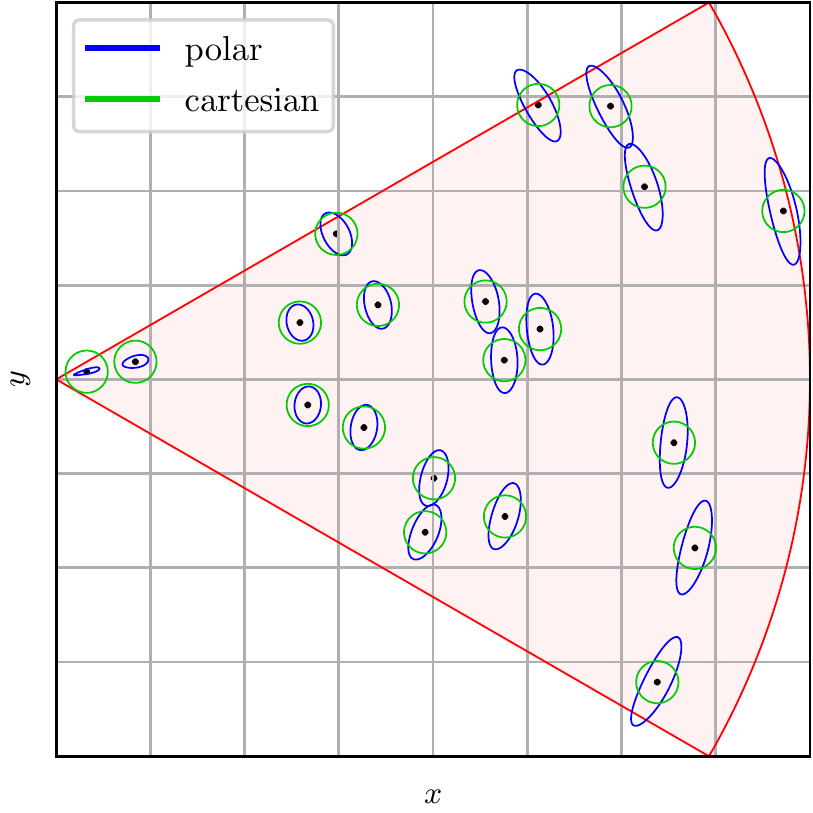}
\caption{Comparison of simulated 3-$\sigma$ error ellipses for polar and cartesian radar measurement models. Cartesian error ellipses tend to overestimate uncertainty at close range and underestimate uncertainty at long range.}
\label{fig:cartesian_vs_polar_ellipses}
\end{figure}

%% file: measurement_model/measurement_model.tex
\section{Polar Measurement Model}

\begin{figure*}
\vspace*{0.15cm}
     \centering
     \begin{subfigure}[h]{0.4\textwidth}
         \centering
         \includegraphics[width=\textwidth]{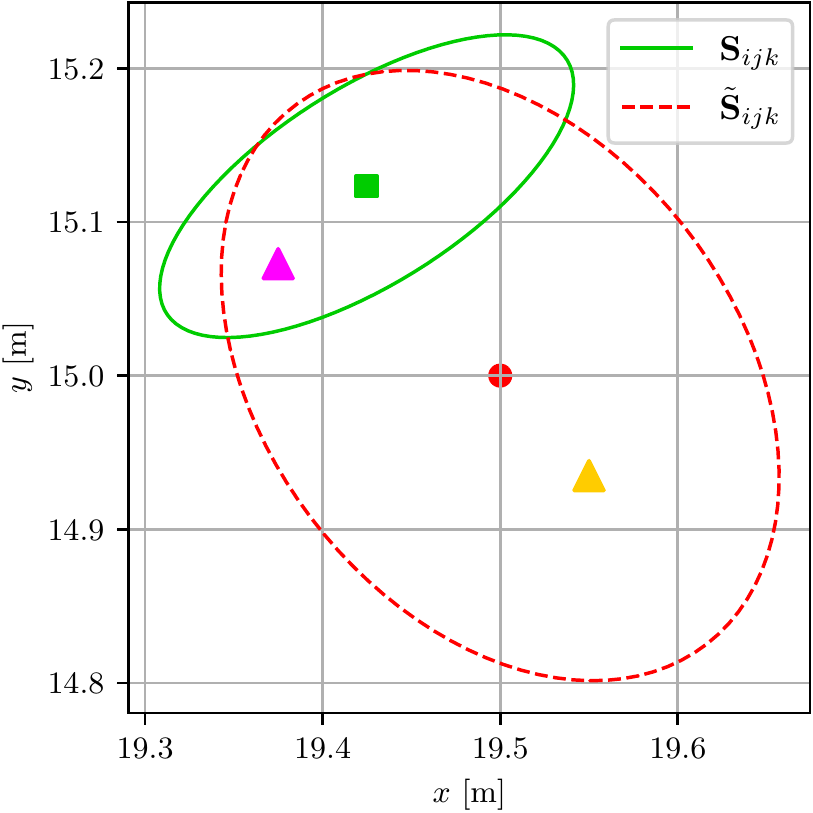}
     \end{subfigure}
     \hspace{7em}%
     \begin{subfigure}[h]{0.4\textwidth}
         \centering
         \includegraphics[width=\linewidth]{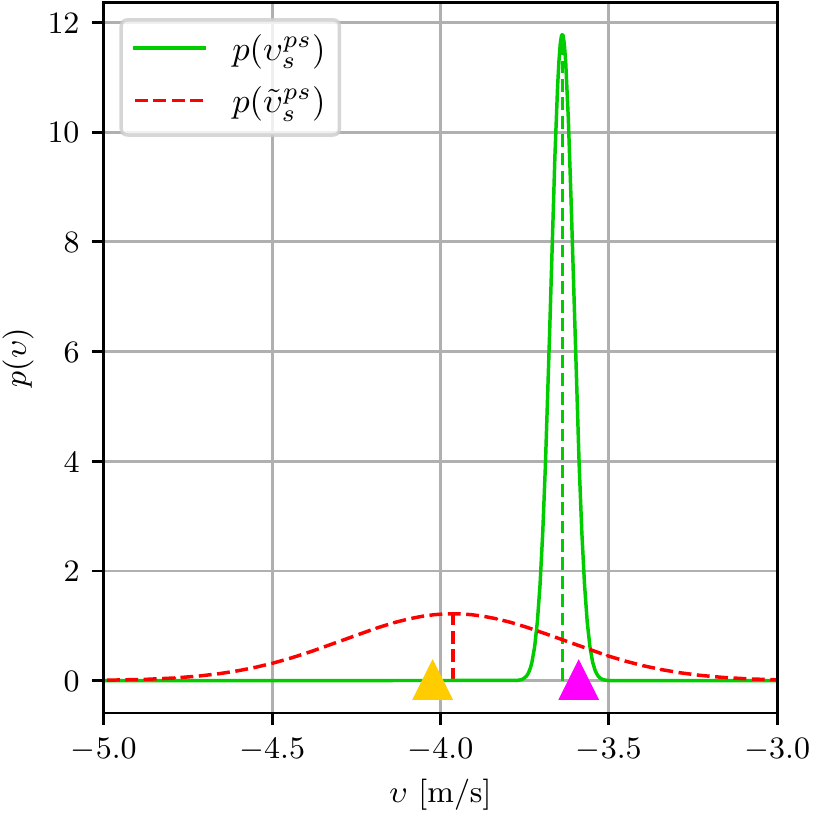}
     \end{subfigure}
     \caption{Inflation of marginal likelihood covariance ellipses (left) and marginal radial velocity distributions (right) due to unmodeled angular acceleration. As $\tilde{\mathbf{S}}_{ijk}$ and $p\left(\tilde{\upsilon}_s^{ps}\right)$ ignore acceleration, they are biased and inflated relative to $\mathbf{S}_{ijk}$ and $p\left(\upsilon_s^{ps}\right)$. As a consequence, the association algorithm selects the incorrect target (orange) rather than the correct target (pink).}
	\label{fig:covariance_inflation}
\end{figure*}
In this section we define a radar measurement model in polar coordinates, which better captures the uncertainties observed from radar detections. We begin by defining the measurement equations and then present the linearization of the proposed measurement model, which we require for our optimization problem. We refer the reader to Section V of \cite{Retan2021} for a detailed discussion of our Cartesian coordinate measurement model.

\subsection{Polar Coordinate Measurement Equations}

Our sensor configuration is depicted in Fig. 2 of \cite{Retan2021}. The radar sensor observes a homogeneous point $\mathbf{p}_j^i$ from a map $\boldsymbol{\mathcal{M}}$. The transformation of $\mathbf{p}_j^i$ from the world frame $\undervec{\boldsymbol{\mathcal{F}}}_i$ to the sensor frame $\undervec{\boldsymbol{\mathcal{F}}}_s$, given by 
\begin{equation}
\mathbf{p}_j^i =
\begin{bmatrix}
\mathbf{r}_s^{p_js} \\
1
\end{bmatrix}
= \mathbf{T}_{sv} \mathbf{T}_{vi} \mathbf{p}_j^i.
\end{equation}
We then defined a simple position measurement in Cartesian coordinates
\begin{align}
\mathbf{g}_{\mathbf{r}}\left( \mathbf{x}_k, \mathbf{p}_j^i, \mathbf{n}_{\mathbf{r},k} \right) &= \mathbf{D} \mathbf{T}_{sv} \mathbf{T}_{v_ki} \mathbf{p}_j^i + \mathbf{n}_{\mathbf{r}, k} \\
\mathbf{n}_{\mathbf{r}, k} &\sim \mathcal{N} \left( \mathbf{0}, \mathbf{R}_{\mathbf{r},k} \right).
\end{align}
In reality, the radar detections are give in polar coordinates, as are the associated measurement uncertainties. Using a measurement covariance defined in Cartesian coordinates is particularly problematic for the angular dimensions, as the model will be overconfident for far-range detections and underconfident for close-range detections. This phenomenon is depicted in Fig. \ref{fig:cartesian_vs_polar_ellipses}, which shows error ellipses from both models for a simulated radar scan.

We therefore propose the use of polar measurement coordinates $\boldsymbol{\rho}_s^{p_js} = \left[r_j \; \theta_j \; \phi_j \; \right]$, where $r$, $\theta$ and $\phi$ denote the range, azimuth and elevation measurements respectively. Our radar detection becomes
\begin{equation}
\mathbf{z}_k = 
\begin{bmatrix}
\boldsymbol{\rho}_s^{p_js} \\
\upsilon_s^{p_js}
\end{bmatrix}.
\end{equation}
where $\upsilon_s^{p_js}$ denotes the radial velocity measurement, with its corresponding measurement model given by (53) from \cite{Retan2021}. The transformation from polar to Cartesian coordinates $s(\cdot)$ is given by
\begin{equation}
\boldsymbol{\rho}_s^{p_js} = \mathbf{s}\left( \mathbf{r}_s^{p_js} \right) = 
\begin{bmatrix}
\left\Vert \mathbf{r}_{s}^{p_{j}s}\right\Vert \\
\tan^{-1}\left(y_{j}/x_{j}\right) \\
\sin^{-1}\left(z_{j}/\left\Vert \mathbf{r}_{s}^{p_{j}s}\right\Vert \right)
\end{bmatrix}.
\end{equation}
We include additive Gaussian noise to obtain the measurement model
\begin{align}
\mathbf{g}_{\boldsymbol{\rho}}\left(\mathbf{x}_k, \mathbf{p}_j^i,  \mathbf{n}_{\mathbf{r},k} \right) &= \mathbf{s}\left(\mathbf{r}_s^{p_js} \right) + \mathbf{n}_{\boldsymbol{\rho},k} \label{eq:polar_measurement_model} \\
\mathbf{n}_{\boldsymbol{\rho}} &\sim \mathcal{N}\left(\mathbf{0}, \mathbf{R}_{\boldsymbol{\rho},k} \right)
, 
\end{align}
which we use to define our radar detection error term
\begin{equation}
\mathbf{e}_{z,ijk} = \mathbf{z}_{ik} - \mathbf{g}\left(\mathbf{x}_k , \mathbf{p}_{ij} \right).
\end{equation}
\subsection{Linearization}
We must now linearize (\ref{eq:polar_measurement_model}) for use in our sliding window optimization problem. In \cite{Retan2021}, we defined the measurement model Jacobian to be 
\begin{equation}
\mathbf{G}=
\left[\begin{array}{cc}
\mathbf{G}_{\mathbf{T}_{vi}}^{\mathbf{r}_{s}^{ps}} & \mathbf{G}_{\boldsymbol{\varpi}_{v}^{iv}}^{\mathbf{r}_{s}^{ps}}\\
\mathbf{G}_{\mathbf{T}_{vi}}^{\upsilon_{s}^{ps}} & \mathbf{G}_{\boldsymbol{\varpi}_{v}^{iv}}^{\upsilon_{s}^{ps}}
\end{array}\right].
\end{equation}
The proposed modification affects only the upper left block $\mathbf{G}_{\mathbf{T}_{vi}}^{\mathbf{r}_{s}^{ps}}$, which now becomes 
\begin{equation}
\mathbf{G}_{\mathbf{T}_{vi}}^{\boldsymbol{\rho}_{s}^{ps}} = \underbrace{\left.\frac{\partial\mathbf{s}\left(\mathbf{r}_{s}^{ps}\right)}{\partial\mathbf{r}_{s}^{ps}}\right|_{\bar{\mathbf{r}}_{s}^{ps}}}_{\mathbf{S}_{\mathbf{r}_{s}^{ps}}^{\boldsymbol{\rho}_{s}^{ps}}} \underbrace{\left.\frac{\partial\mathbf{r}_{s}^{ps}\left(\mathbf{T}_{vi}\right)}{\partial\boldsymbol{\epsilon}}\right|_{\bar{\mathbf{T}}_{vi}}}_{\mathbf{G}_{\mathbf{T}_{vi}}^{\mathbf{r}_{s}^{ps}}},
\end{equation}
where $\mathbf{S}_{\mathbf{r}_{s}^{ps}}^{\boldsymbol{\rho}_{s}^{ps}}$ is the Jacobian of the cartesian-to-polar coordinate transformation.

%% file: data_association/data_association.tex
\section{Data Association}
\begin{table*}[htb]\centering
\vspace*{0.15cm}
	\begin{tabular}{c c L L L L L L L} \hline
		\multirow{2}{*}{Seq. No.}	& \multirow{2}{*}{Distance (km)}	& \multicolumn{2}{c}{Trans. Error $SE(3)$ (\%)}	& \multicolumn{2}{c}{Rot. Error $SE(3)$ (deg/m)}	& \multicolumn{2}{c}{Runtime (ms)}					\\ 
     	                            &                        			& CV/C		& CA/P              				& CV/C		& CA/P									& CV/C		& CA/P		\\ \hline
        1     						& 2.74 								& 1.9503	& \textbf{1.1624}  					& 0.0088	& \textbf{0.0054}						& 11.785	& \textbf{10.756}	\\
        2     						& 3.13	  							& 2.7746	& \textbf{1.5264}					& 0.0144	& \textbf{0.0081}						& 13.908	& \textbf{12.856}   \\
		3     						& 4.91  							& 2.8433	& \textbf{1.6398}					& 0.0119	& \textbf{0.0088}						& \textbf{10.522} & 11.453	\\
		4     						& 4.93					  			& 2.4284	& \textbf{1.7471}   				& 0.0140	& \textbf{0.0090}						& 11.270	& \textbf{10.872}   \\
        5     						& 6.66  							& 2.4353    & \textbf{1.4779}   				& 0.0113    & \textbf{0.0079}						& 13.079	& \textbf{11.526}	\\ 		
        6     						& 10.51 							& 1.9255	& \textbf{1.4824}					& 0.0084    & \textbf{0.0075}						& 11.762	& \textbf{11.550}   \\
        7					   		& 10.94 							& 2.2557	& \textbf{1.7214}					& 0.0094	& \textbf{0.0076}						& 11.8861	& \textbf{11.025}	\\
        8     						& 11.61 							& 2.4296	& \textbf{1.8293}					& 0.0113    & \textbf{0.0081}						& 13.183	& \textbf{12.128}	\\
        9     						& 14.85 							& 3.1619	& \textbf{1.9191}					& 0.0127    & \textbf{0.0082}						& 12.501	& \textbf{11.941}   \\ \hline
        overall						& 70.28								& 2.5079	& \textbf{1.6878} 					& 0.0111 	& \textbf{0.0079}						& 12.272	& \textbf{11.617}	\\ \hline

	\end{tabular}
	\caption{Odometry estimation results. Best results are marked in bold.}
\label{tab:results}
\end{table*}  
To estimate motion well, we must be able to robustly associate radar detections from a radar scan $\boldsymbol{\mathcal{Z}}_k$ to points in the map $\boldsymbol{\mathcal{M}}$. In our previous work, we used a maximum likelihood data association approach based on the marginal distribution 
\begin{multline}
	p(\mathbf{z}_{\theta k} \vert \mathbf{p}_j, \boldsymbol{\mathcal{Z}}_{1:k-1}) = \\
	 \dfrac{1}{\sqrt{(2\pi)^4\vert\mathbf{S}_{\theta jk}\vert}}\exp\left(-\frac{1}{2}\mathbf{e}_{\mathbf{z}, \theta jk}^T\mathbf{S}_{\theta jk}^{-1}\mathbf{e}_{\mathbf{z}, \theta jk})\right), \label{eq:marginal_likelihood}
\end{multline}
where $\mathbf{S}_{\theta jk}$ is the marginal likelihood covariance given by
\begin{equation}
	\mathbf{S}_{\theta jk} = \mathbf{G}_{jk} \mathbf{P}_k \mathbf{G}_{jk}^T + \mathbf{R}_{\theta k}.
\end{equation}
and the index $\theta$ represents the radar detection assignment $\theta_k(j)$ of the $j$th map point. We then take the negative log likelihood of this expression and minimize with respect to the measurement index $i$: 
\begin{equation}
	\hat{\theta}_k(j) = \argmin_{i, \forall i \in \{1,\dots,\vert\boldsymbol{\mathcal{Z}}_k\vert\}}\mathbf{e}_{\mathbf{z}, ijk}^T \mathbf{S}_{ijk}^{-1} \mathbf{e}_{\mathbf{z}, ijk} = d_{ij}^2.
\end{equation}
We refer the reader to \cite{Retan2021} Section VI for a more thorough discussion of our data association process. Here we investigate the effect of ignoring acceleration, meaning $\mathbf{x}_k = \left\{ \mathbf{T}_k, \boldsymbol{\varpi}_k, \mathbf{0} \right\}$ -- as is the case with a constant velocity motion prior -- on the data association process. We define this measurement error between the $i$th detection and the $j$th map point to be
\begin{equation}
\mathbf{e}_{\mathbf{z}, ijk} = \mathbf{z}_{ik} - \mathbf{g}\left(\boldsymbol{\gamma}_k \oplus \check{\mathbf{x}}_{k}, \mathbf{p}_j^i, \mathbf{n}_{jk} \right),
\end{equation}
where $\boldsymbol{\gamma}_k \oplus \check{\mathbf{x}}_k$ denotes the predicted state and $\mathbf{n}_{jk}$ the measurement noise for the $j$th map point. We can use a first order Taylor expansion to approximate the error as 
\begin{align}
\mathbf{e}_{\mathbf{z}, ijk} &\approx \mathbf{z}_{ik} - \mathbf{g}\left( \check{\mathbf{x}}_{k}, \mathbf{p}_j^i, \mathbf{0} \right) - \mathbf{G}_{jk} \boldsymbol{\gamma}_k + \mathbf{n}_{jk} \\
&= \bar{\mathbf{e}}_{\mathbf{z}, ijk} - \mathbf{G}_{jk} \boldsymbol{\gamma}_k + \mathbf{n}_{jk}.
\end{align}
If the ego vehicle accelerated between $t_{k-1}$ and $t_{k}$ we will have $E[\mathbf{e}_{ijk}] \neq \mathbf{0},$
resulting in a biased state estimates, which was demonstrated in \cite{Tang2018}. Here we consider the effect of unmodeled acceleration on the covariance $\mathbf{S}_{ijk}$, which can be written as
\begin{align*}
\mathbf{S}_{ijk} &= E[\mathbf{e}_{\mathbf{z},ijk} \mathbf{e}_{\mathbf{z},ijk}^T] \\
&= E[\left( \bar{\mathbf{e}}_{\mathbf{z}, ijk} - \mathbf{G}_{jk} \boldsymbol{\gamma}_k + \mathbf{n}_k \right) \left( \bar{\mathbf{e}}_{\mathbf{z}, ijk} - \mathbf{G}_{jk} \boldsymbol{\gamma}_k + \mathbf{n}_k \right)^T] \\
&= E[\bar{\mathbf{e}}_{\mathbf{z},ijk} \bar{\mathbf{e}}_{\mathbf{z},ijk}^T] + \mathbf{G}_{jk} \mathbf{P}_k \mathbf{G}_{jk}^T + \mathbf{R}_{ik}. \numberthis
\end{align*}
We see that $\mathbf{S}_{ijk}$, which is used for calculating the Mahalanobis distance in data association, is inflated by the outer product of the residual. The consequence is unnecessarily large association ellipses, which greatly increases the frequency of incorrect associations. Fig. \ref{fig:covariance_inflation} depicts both covariance ellipses and the marginal radial velocity distributions associated with the distribution (\ref{eq:marginal_likelihood}). The ellipse generated by $\tilde{\mathbf{S}}_{ijk}$ and distribution $p\left(\tilde{\upsilon}_{s}^{p_js}\right)$ are biased by ignoring acceleration, while $\mathbf{S}_{ijk}$ and $p\left(\upsilon_{s}^{p_js}\right)$ use the full state estimate.

%% file: evaluation/evaluation.tex
\section{Evaluation}
\begin{figure}[t]
\vspace*{0.15cm}
     \centering
     \begin{subfigure}[t]{0.45\textwidth}
         \centering
         \includegraphics[width=\textwidth, trim=0 25 0 55, clip]{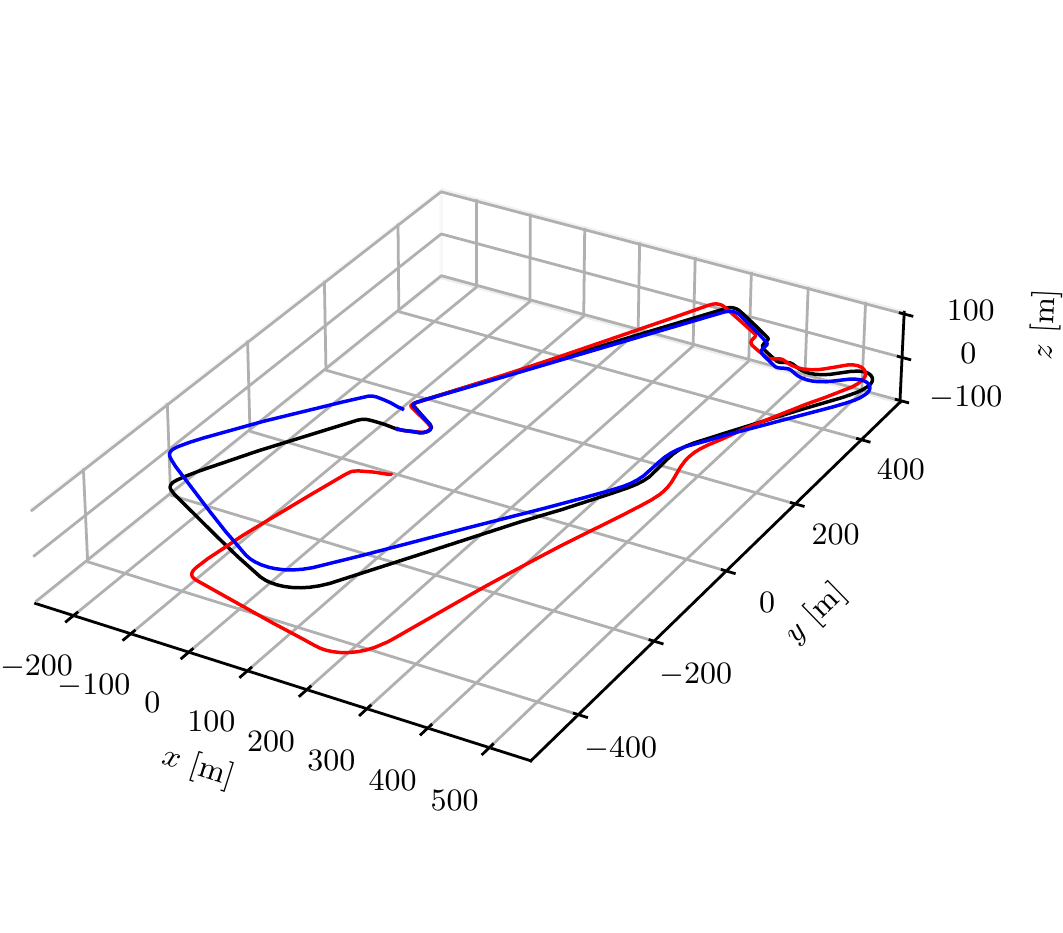}
     \end{subfigure}
     \hfill
     \centering
     \begin{subfigure}[t]{0.45\textwidth}
         \centering
         \includegraphics[width=\textwidth]{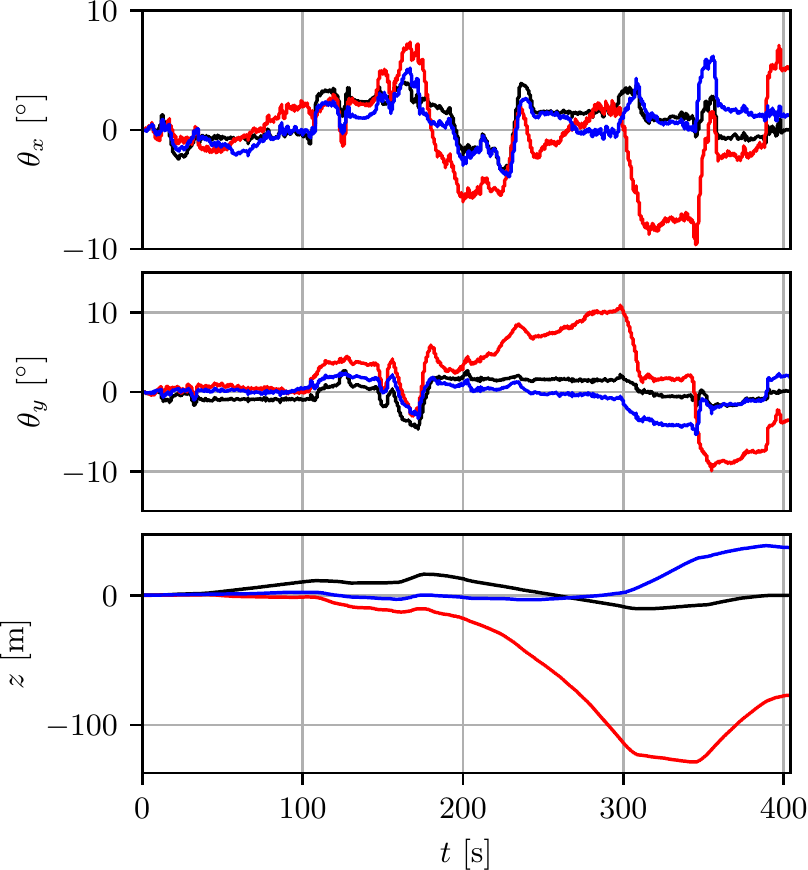}
     \end{subfigure}
     \caption{Odometry results from Friedrichshafen Exhibition Center: ground truth (black), constant velocity with Cartesian measurement model (red), constant acceleration with polar measurement model (blue).}
\label{fig:bias_reduction}
\end{figure}
We evaluate our approach using data from a front-facing prototype 77 GHz FMCW MIMO imaging radar mounted on a light commercial vehicle with a 13 Hz frame rate. The sensor has an accuracy of up to 0.15° in azimuth, 0.3° in elevation, 0.01 m/s in radial velocity and 2 cm in range. Exact values depend on the characteristics of the target and the sensor operation mode, as resolution is sacrificed for range at high driving speeds. The vehicle drives different routes throughout Friedrichshafen, Germany and the surrounding Bodensee region. In total we capture over 3 hours of measurements amounting to around 70 driven kilometers. The scenes are very diverse and challenging in their makeup, ranging from city centers to highways with speeds exceeding 90 km/h.

For evaluation we use a KITTI-style metric \cite{6248074}, which calculates translational and rotational errors at distances of 100, 200,...,800 meters, normalized by distance. We compare our constant acceleration / polar measurement model (CA/P) presented here to our previous constant velocity / Cartesian measurement model (CV/C). The relevant hyperparameters for both models are the power spectral density matrices $\mathbf{Q}$ of the constant acceleration and constant velocity models motion models, as well as the measurement noise covariance matrices $\mathbf{R}$ of the Cartesian and polar radar point models. These are tuned for maximum performance for each approach using measurement sequence 1.

From our previous work we determined that the size of the map $\boldsymbol{\mathcal{M}}$ can significantly influence performance. Here we hold the map size constant at 1500 points for both models. We set the association threshold parameter $\gamma$ to the 95\% confidence interval of the  marginal likelihood distribution.

%% file: results/results.tex
\section{Results}
Results from our evaluation can be found in Table \ref{tab:results}. For the CA/P model we see a consistent and substantial improvement in odometry performance across all measurement sequences. These sequences vary in terms of the scene complexity and driving behavior. In some sequences, the CV/C model becomes unstable due to difficult traffic conditions such as complete sensor blockage from a vehicle in front of the sensor while driving at low speeds. This is precisely the situation where the lack of acceleration in the vehicle state becomes problematic for data association. The inflated marginal likelihood ellipse causes associations with dynamic objects in front of the vehicle which occlude its field of view. In contrast, the CA/P is much more robust to these situations and runs stably over all measurement sequences.

Interestingly, we see that the constant acceleration model has a faster runtime than the constant velocity model, despite the larger state dimension and overall more complex motion prior calculation. Again, this can be attributed to smaller marginal likelihood ellipses for data association, which speed up the nearest neighbor search in our particular KD-tree implementation.

In Fig. \ref{fig:bias_reduction} we see that our new approach significantly reduces drift in pitch, roll and altitude -- precisely the dimensions which were so problematic in \cite{Retan2021} and were observed to be a limitation of constant velocity priors in \cite{Tang2018}. The improvement seems to be even more pronounced here, which could be attributed to the radial velocity measurement and its sensitivity to accurate velocity priors. We speculate that the mounting position at the front of the vehicle makes the estimate much more sensitive to motion priors and calibration effects due to the large distance from the vehicles center of rotation, which is generally near the center of the rear axle.

We note that the proposed method is very sensitive to parameterization of the matrices $\mathbf{R}$ and $\mathbf{Q}$. Data association becomes precise only when these quantities describe the underlying physical processes well. Future work in this area should focus on the optimizing noise covariances in a principled way as in \cite{Caesar2019-as}.

\addtolength{\textheight}{-12cm}

%% file: conclusion/conclusion.tex
\section{Conclusion}
In this paper we demonstrated an approach to radar odometry using a constant acceleration motion prior and polar measurement model. Our main contribution was the derivation of a fast and accurate constant acceleration motion prior based on the Magnus expansion. Additionally, we analyze the importance of acceleration for the data association process. Our estimation results are substantially more accurate than those achieved with a constant velocity motion prior and Cartesian measurement model. Future work should extend this approach to 3D point-cloud based radar localization.